\documentclass{article}

\PassOptionsToPackage{numbers,square}{natbib}
\usepackage[preprint]{neurips_2026}

\usepackage[utf8]{inputenc}
\usepackage[T1]{fontenc}
\usepackage{float}
\usepackage{hyperref}
\usepackage{url}
\usepackage{booktabs}
\usepackage{multirow}
\usepackage{tabularx}
\usepackage{amsmath}
\usepackage{amssymb}
\usepackage{amsfonts}
\usepackage{nicefrac}
\usepackage{microtype}
\usepackage{graphicx}
\usepackage{xcolor}
\usepackage{enumitem}
\setlist{leftmargin=*}

\newcommand{\densefourx}{Dense-4x}
\newcommand{\posdelta}[1]{$_{\textcolor{green!50!black}{\scriptscriptstyle +#1}}$}
\newcommand{\negdelta}[1]{$_{\textcolor{red!70!black}{\scriptscriptstyle -#1}}$}

\newcommand{\blfootnote}[1]{
  \begingroup
  \renewcommand{\thefootnote}{}
  \renewcommand{\theHfootnote}{blfootnote.\arabic{footnote}}
  \begin{NoHyper}\footnotetext{#1}\end{NoHyper}
  \addtocounter{footnote}{-1}
  \endgroup
}

\title{Continual LLM Upcycling: A Predictor-Gated Bank-Wise Sparsity Training Recipe for Dense-to-Sparse LLMs}

\author{
  \begin{tabular}{c}
    Ruixuan Huang\textsuperscript{1} \quad
    Jinyuan Shi\textsuperscript{2} \quad
    Hantao Huang\textsuperscript{3,*} \quad
    Yifan Huang\textsuperscript{3}
    \\[2pt]
    Ziyi Guan\textsuperscript{3} \quad
    Hao Zeng\textsuperscript{2} \quad
    Ian En-Hsu Yen\textsuperscript{2,*} \quad
    Minghui Yu\textsuperscript{3,*}
  \end{tabular}
}

\begin{document}

\maketitle
\blfootnote{\textsuperscript{1}HKUST. \textsuperscript{2}Moffett AI. \textsuperscript{3}Independent Contributor. \textsuperscript{*}Corresponding authors.}

\begin{abstract}
We study dense-to-sparse continual training as a way to construct channel-sparse large language models from dense checkpoints. Starting from a Qwen2.5-8B dense backbone, we continue training at 32K context and introduce a predictor-gated sparse SwiGLU FFN in the 32K stage. For each token and layer, we use a low-rank predictor to produce FFN-channel routing logits. We then apply a bank-wise top-$k$ rule to retain 16 channels in every 64-channel bank, yielding 4x sparsity in the FFN intermediate activation. Unlike post-hoc sparse inference methods, the routing module is placed on the main language modeling path and optimized during continual training, enabling the dense model to be upcycled into a hardware-oriented sparse model. We report the architecture, training recipe, benchmark performance, and training lessons. We also identify a layer-local long-context failure mode on RULER-CWE and propose a single-layer repair algorithm that substantially improves the affected length range.
\end{abstract}

\section{Introduction}

The cost of large language models (LLMs) is increasingly dominated by the parameter size and the amount of computation for every token. This is especially apparent in decoder-only Transformers \citep{vaswani2017attention}, where each token passes through a dense feed-forward network (FFN) in every layer. In modern SwiGLU-based models \citep{shazeer2020glu}, the FFN expands the hidden state to a much larger intermediate dimension, applies element-wise gating, and projects the result back to the model dimension. During long-context training and long-prompt prefill, this computation is applied over large token blocks, even though only a subset of intermediate channels are important for any particular token \citep{liu2023dejavu}. Reducing this token-wise FFN cost is therefore a central requirement for making long-context LLMs more efficient.

Conditional computation offers a natural route to this goal. Mixture-of-Experts (MoE) models reduce computation by routing each token to a small number of expert FFNs, but they introduce separate expert parameter pools, expert-level routing, load-balancing concerns, and often all-to-all communication \citep{lepikhin2020gshard,fedus2022switch}. A finer-grained alternative is to view the intermediate channels of a dense FFN as neuron-level computational units and activate only a subset of them for each token. It preserves the original dense FFN parameterization while exposing conditional computation inside each layer and can turn a dense FFN into a sparse sub-FFN whose active components change with the token.

However, not all channel sparsity leads to usable acceleration. Contextual sparsity methods demonstrate that LLM computation is strongly input-dependent \citep{liu2023dejavu,akhauri2024shadowllm,zhou2024sirius}. Channel-level methods \citep{wu2025mixturechannels} show that many FFN channels can be suppressed on a per-token basis. However, if the sparse mask is derived from native FFN activations, the routing signal may arrive too late to skip the dominant gate and up projections. For channel sparsity to become a hardware-facing computation path, the sparse mask must be available before the main FFN channel computation is materialized.

This paper studies the dense-to-sparse continual-training process, seeking to convert a dense LLM into a channel-wise sparse model. We introduce a lightweight low-rank predictor that maps each token hidden state to logits over FFN intermediate channels. The predictor produces the routing signal before the FFN computations. We then impose a fixed bank-wise top-$k$ rule, which keeps 16 active channels in every 64-channel bank, giving a 4$\times$ reduction in active FFN intermediate width. The resulting sparse path is trained with hard sparse execution in the forward pass and a soft surrogate gradient through the top-$k$ boundary. We instantiate this idea in a Qwen2.5-8B model backbone \citep{yang2024qwen25}. The model is first trained from scratch at 8K context length and then continued at 32K context length. Sparsity is introduced late in the 32K stage, reflecting a practical upcycling scenario \citep{komatsuzaki2022sparseupcycling}.

Our experiments compare the resulting sparse model with (1) a dense companion run and (2) a naive 4$\times$ sparse baseline based on dense activation magnitude (\densefourx{}). Our sparse model is consistently closer to the dense baseline than the naive \densefourx{} baseline across a broad set of evaluation benchmarks, indicating that channel sparsity can be trained as part of the language-modeling path rather than applied only as a post-training compression heuristic.

This paper reveals that sparse upcycling is not lossless. We highlight two instructive failure modes during our training process. First, channel-level routing does not inherit MoE-style balancing in a straightforward way. Second, long-context evaluation exposes failures that are nearly invisible in broad short-context benchmarks. On RULER-CWE \citep{hsieh2024ruler}, the sparse model remains healthy at shorter lengths but develops a sharp 12K--16K cliff. We propose a dense fallback intervention that localizes much of this failure to a single layer and the final question-answer suffix.

\paragraph{Contributions.}
In summary, this paper makes the following contributions:
\begin{itemize}
    \item \textbf{Predictor-gated channel sparsity pattern.} We introduce a predictor-gated sparse pattern in which a low-rank token-wise predictor produces channel routing logits before FFN channel computation, and a hard bank-wise top-$k$ rule retains 16 channels in every 64-channel bank. It can skip inactive gate, up, and down channels when supported by specialized kernels or sparse hardware.
    \item \textbf{Dense-to-sparse continual training recipe.} We present a dense-to-sparse continual training recipe that converts a Qwen2.5-8B dense model trained at 8K and continued at 32K into a predictor-gated 4$\times$ sparse FFN model, with limited performance degradation.
    \item \textbf{Training lessons.} We report two training lessons. First, MoE-style balance bias does not transfer cleanly to bank-wise channel routing. Second, RULER-CWE reveals a layer-local long-context failure that can be partially repaired through a single-layer intervention.
\end{itemize}

\section{Related work}

\paragraph{Contextual sparsity.}
Prior work shows that LLM computation is input-dependent and can often be sparsified. Dejavu predicts token-dependent sparse sets of attention heads and MLP neurons for efficient inference \citep{liu2023dejavu}, while ShadowLLM improves predictor-based contextual sparsity with stronger importance criteria and deployment-oriented predictors \citep{akhauri2024shadowllm}. Sirius highlights that training-free contextual sparsity can be fragile on complex generation tasks and introduces correction mechanisms to recover quality \citep{zhou2024sirius}. This paper is more training-centric, where the sparse gate is inserted into the main SwiGLU FFN computation and optimized during continual training.

\paragraph{Channel-sparse FFNs.}
Mixture-of-Channels (MoC) introduces persistent channel sparsity inside SwiGLU FFNs by selecting the top-$k$ channels from the native gate projection for each token \citep{wu2025mixturechannels}. However, the sparse decision is still derived from the native gate output, so the gate projection must be evaluated before the active set is known. SparkTransformer proposes a new sparse architecture for both FFN and attention, using top-$k$ activation masks and low-cost predictors formed by reallocating existing dimensions \citep{you2025spark}. This predictor is an internal consequence of Spark's redesigned parameterization. The resulting sparse design is not directly compatible with an existing SwiGLU architecture.

\paragraph{MoE and dense-to-sparse conversion.}
Expert-level conditional computation routes tokens to a small number of full FFN experts. Related dense-to-sparse approaches include MoEfication, which partitions FFN parameters into functional experts and builds expert routers \citep{zhang2022moefication}, and Aran et al., which initializes sparse MoE models from dense checkpoints \citep{komatsuzaki2022sparseupcycling}. The channel-level view we use here is MoE-like in spirit because FFN intermediate channels behave as fine-grained conditional units, but it does not introduce independent expert parameter pools or expert-level all-to-all dispatch. It preserves the dense SwiGLU parameterization and performs token-wise channel selection.

\begin{figure}[t]
    \centering
    \includegraphics[width=\linewidth]{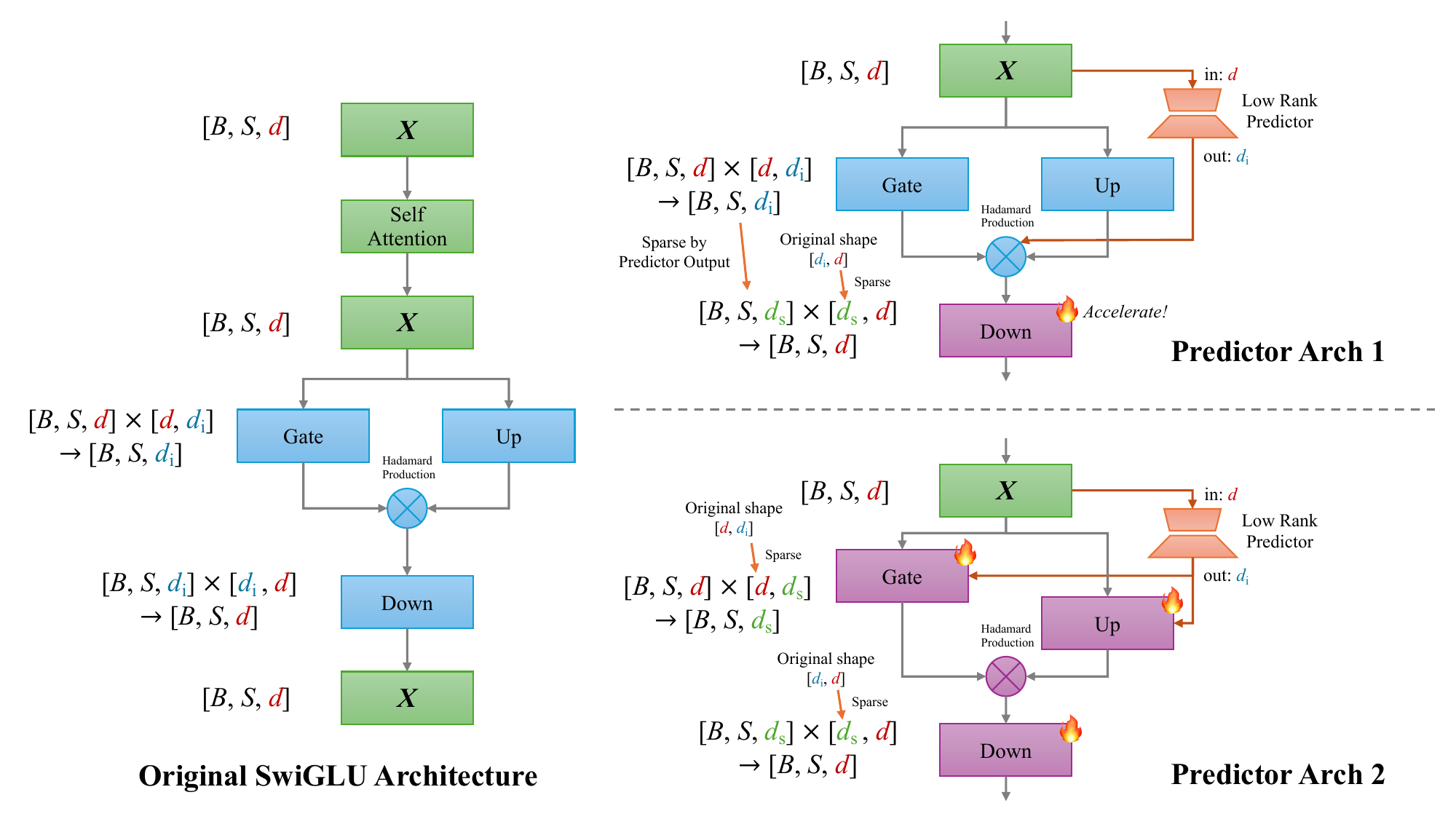}
    \caption{Our proposed predictor-gated sparse SwiGLU FFN. A low-rank predictor maps the token hidden state to FFN-channel routing logits.}
    \label{fig:swiglu_arch}
\end{figure}

\section{Methodology}
\label{sec:method}

This section describes the predictor-gated sparse FFN that we use in the dense-to-sparse continual-training process. We start from the dense SwiGLU computation, then introduce the proposed low-rank predictor, the bank-wise top-$k$ routing, and technical details that keep the sparse path trainable.

\subsection{Dense SwiGLU FFN}

For layer $l$ and token position $t$, let the FFN input be $x_{l,t} \in \mathbb{R}^{d}$, where $d$ is the hidden dimension. We use $d_i$ for the dense FFN intermediate dimension. A standard SwiGLU FFN computes
\begin{align}
u_{l,t} &= W^g_l x_{l,t}, &
v_{l,t} &= W^u_l x_{l,t}, \\
m^{\mathrm{dense}}_{l,t} &= \mathrm{SiLU}(u_{l,t}) \odot v_{l,t}
                         = u_{l,t} \odot \sigma(u_{l,t}) \odot v_{l,t}, \\
y^{\mathrm{dense}}_{l,t} &= W^o_l m^{\mathrm{dense}}_{l,t}.
\end{align}
where $u_{l,t}, v_{l,t}, m^{\mathrm{dense}}_{l,t} \in \mathbb{R}^{d_i}$. In our 8B-scale training configuration, $d=4096$ and $d_i=14336$.

\subsection{Predictor-gated Channel Sparsity}

We introduce a per-layer predictor $P_l$ that maps the token hidden state to logits over FFN intermediate channels:
\begin{align}
p_{l,t} = P_l(x_{l,t}) \in \mathbb{R}^{d_i}.
\end{align}
The sparse gate $g_{l,t}$ is built from these logits and replaces the dense SwiGLU gate $\sigma(u_{l,t})$:
\begin{align}
m_{l,t} &= u_{l,t} \odot g_{l,t} \odot v_{l,t}, \\
y_{l,t} &= W^o_l m_{l,t}.
\end{align}
Thus, the predictor directly modulates the FFN intermediate activation and receives gradients from the language modeling loss.

The predictor is a small two-layer low-rank module
\begin{align}
P_l(x) = W^{p,2}_l W^{p,1}_l x,
\end{align}
with dimensions
\begin{align}
W^{p,1}_l &: \mathbb{R}^{d} \rightarrow \mathbb{R}^{448}, &
W^{p,2}_l &: \mathbb{R}^{448} \rightarrow \mathbb{R}^{d_i}.
\end{align}
The hidden size 448 is $d_i/32$. The predictor output also includes a bank-level affine calibration with learnable scale and bias parameters shared within each 64-channel bank.

\subsection{Bank-wise Hard Top-$k$ with Soft Surrogate Gradients}

The FFN intermediate dimension is partitioned into fixed banks of size $C_{\mathrm{bank}}=64$. For each token, layer, and bank, we keep $k=16$ channels:
\begin{align}
p_{l,t} &= [p^{(1)}_{l,t}, \ldots, p^{(G)}_{l,t}], \quad p^{(b)}_{l,t} \in \mathbb{R}^{C_{\mathrm{bank}}}, \\
S^{(b)}_{l,t} &= \operatorname{topk}(p^{(b)}_{l,t}, k=16).
\end{align}
The gate is
\begin{align}
g_{l,t,i} =
\begin{cases}
\sigma(p_{l,t,i}), & i \in S^{(b)}_{l,t}, \\
0, & \text{otherwise}.
\end{cases}
\end{align}
With $d_i=14336$, this gives $G=d_i/C_{\mathrm{bank}}=224$ banks and keeps $d_s=Gk=3584$ channels per token per layer. Here, $d_s$ is the sparse intermediate dimension shown in Figure~\ref{fig:swiglu_arch}.

Adding a predictor and a discrete top-$k$ operator changes the FFN routing rule, but the end-to-end training path remains a standard language-modeling computation graph with a custom gradient for the mask. In the forward pass, each bank first normalizes predictor logits, selects the top-16 channels, and constructs a binary mask $h_{l,t}$. The FFN intermediate activation is then
\begin{align}
m_{l,t} = u_{l,t} \odot v_{l,t} \odot \sigma(p_{l,t}) \odot h_{l,t},
\end{align}
so inactive channels are exactly zero in the model path seen by the loss.

In the backward pass, gradients through $W^o_l$, $W^g_l$, $W^u_l$, and the selected intermediate activations follow ordinary automatic differentiation under the chosen mask. The non-differentiable top-$k$ selection is handled by a soft top-$k$ surrogate: around each bank's selection threshold, the hard indicator is assigned a sigmoid-like local gradient. Thus the language-modeling loss can still update the predictor logits and predictor parameters near the routing boundary.

\subsection{Hardware Implications}

Our training and validation are performed on GPUs as a functional simulation of bank-wise sparse FFN semantics. The model is trained with a hard sparse gate in the forward path, but the GPU implementation still materializes the dense gate and up projections, $W^g x$ and $W^u x$, before applying the top-$k$ mask. As a result, the practically accelerated part on GPUs is mainly the downstream projection $W^o m$, where the zeroed intermediate activation can be gathered and multiplied against only the selected rows/columns. Our experiments therefore validate sparse semantics and model quality rather than demonstrating measured 4x end-to-end GPU wall-clock speedup.

The algorithmic principle is stronger than what our GPU implementation exposes. For any inactive channel $i$ with $h_{l,t,i}=0$, the term
\begin{align}
u_{l,t,i} \, v_{l,t,i} \, \sigma(p_{l,t,i}) \, h_{l,t,i}
\end{align}
contributes exactly zero to the FFN output. Dedicated sparse hardware can therefore place the predictor-generated mask before the FFN projections and avoid computing inactive channels in all three FFN stages: the gate projection $W^g$, the up projection $W^u$, and the down projection $W^o$. Under the 64:16 bank rule, the ideal sparse execution shape is reduced from the dense intermediate width $d_i$ to the selected sparse width $d_s=d_i/4$ for each token and layer, while preserving the same mathematical sparse FFN path. The intended efficiency target is therefore dedicated sparse hardware or specialized sparse kernels.

\section{Training recipe}

This section reports the training-side context needed to interpret the sparse model and its comparisons. We describe the Qwen2.5-style model backbone and the training data curriculum.

\subsection{Model Backbone}
\label{sec:backbone}

The dense backbone follows the Qwen2.5 architectural family rather than a post-hoc modification of an open-source checkpoint \citep{yang2024qwen25}. The 8B-scale configuration uses 32 layers, hidden size 4096, FFN intermediate dimension 14336, 32 query heads, and 4 shared key-value heads. The tokenizer vocabulary size is 155136.

We also follow the Qwen2.5-style companion recipe. We double the number of query heads in grouped-query attention to increase the parameter count. The sparse model adds a per-layer low-rank predictor that produces token-wise FFN channel routing logits before bank-wise top-$k$ selection. The predictor adds approximately 0.26B trainable parameters.

\subsection{Pretraining Data}

The pretraining corpus is a multilingual, multi-domain mixture designed to transfer strong English capability while progressively increasing Chinese coverage. Table~\ref{tab:stage_data_mixture} summarizes the active 8K and 32K schedules used for the model. The 8K dense stage uses an 8T-token schedule. The 32K continuation uses a separate long-context mixture with 4T tokens. During the 32K continuation, the RoPE base is increased to support the longer sequence length.

\begin{table}[h]
\centering
\caption{Data mixture and context-length curriculum. Content columns report expected sampling shares within each stage; token counts use the reported training budget.}
\label{tab:stage_data_mixture}
\scriptsize
\setlength{\tabcolsep}{2pt}
\resizebox{\linewidth}{!}{
\begin{tabular}{lccccccccccc}
\toprule
Stage & Seq. & Tokens & RoPE base & Chinese & English & Wiki/ref. & Code & Math & Domain & Reasoning & Parallel \\
\midrule
\multirow[t]{3}{*}{Stage 1} & 8K & 0--3T & $10^{4}$ & 0.0\% & 67.3\% & 0.0\% & 21.0\% & 1.5\% & 6.8\% & 3.0\% & 0.4\% \\
 & 8K & 3--5T & $10^{4}$ & 15.3\% & 50.0\% & 0.0\% & 20.0\% & 1.5\% & 8.8\% & 4.0\% & 0.4\% \\
 & 8K & 5--8T & $10^{4}$ & 30.6\% & 32.7\% & 0.0\% & 19.0\% & 1.5\% & 10.8\% & 5.0\% & 0.4\% \\
Stage 2 & 32K & 4T & $10^{6}$ & 7.6\% & 7.6\% & 3.8\% & 33.5\% & 10.2\% & 20.1\% & 17.2\% & 0.0\% \\
\bottomrule
\end{tabular}
\!}
\end{table}

\subsection{Sparse Continual Training}

The sparse branch is introduced late in Stage 2. The dense model first runs the full 8K stage and then continues at 32K. Predictor sparsity is introduced only after roughly 3T tokens. In particular, this sparse branch uses weight decay 0.06, whereas the non-sparse reference recipe uses weight decay 0.1. In Figure~\ref{fig:training_loss_32k}, the x-axis is aligned to the 4T-token 32K stage budget.

We summarize the whole training process as follows:
\begin{enumerate}
    \item \textbf{Dense, 8K stage}: train from scratch with the same Qwen2.5-8B architecture.
    \item \textbf{Dense, 32K stage}: resume from the dense 8K checkpoint, increase the context length to 32K, and continue dense training for around 3T tokens.
    \item \textbf{Predictor warm-up}: add the low-rank predictor and train it as an auxiliary branch that imitates the native gate signal; this branch does not affect the main loss and lasts only a small number of tokens.
    \item \textbf{Predictor-gated dense FFN}: route the main FFN gate through the predictor and sigmoid so that the predictor affects the language-modeling loss, but do not yet apply hard top-$k$ channel selection.
    \item \textbf{Predictor-gated top-$k$ sparse FFN}: add bank-wise top-$k$ selection to impose 4x channel sparsity.
\end{enumerate}

Appendix~\ref{app:predictor_topk_ablation} reports a model diagnostic ablation that motivates this predictor-before-top-$k$ ordering. This ablation runs on a 200M model and is separate from our main evaluation.

\section{Evaluation}

We evaluate the sparse model against two companion baselines:
\begin{itemize}
    \item \textbf{Dense companion}: a fully dense run with comparable data and training budget.
    \item \textbf{\densefourx}: a naive 4x sparse baseline based on activation magnitude, applied post-hoc to the dense companion's training results.
    \item \textbf{Ours}: our predictor-gated channel-wise 4x sparsity.
\end{itemize}

Table~\ref{tab:autoeval_benchmarks_32k} compares the benchmarking performance of the three models. The \densefourx{} and sparse columns append their deltas relative to the dense companion; green indicates a positive delta and red indicates a negative delta. We summarize the evaluated benchmarks as six categories:
\begin{itemize}
    \item \textbf{Factual QA}: SimpleQA \citep{wei2024simpleqa} and Chinese SimpleQA \citep{he2024chinesesimpleqa}.
    \item \textbf{Code}: MBPP \citep{austin2021mbpp}, MBPP+ \citep{liu2023evalplus}, MultiPL-E \citep{cassano2023multiple}, MBXP \citep{athiwaratkun2022mbxp}, and LiveCodeBench \citep{jain2024livecodebench}.
    \item \textbf{Reasoning}: KOR-Bench \citep{ma2024korbench}, ZebraLogic \citep{lin2025zebralogic}, ProcBench \citep{fujisawa2024procbench}, ARC-AGI \citep{chollet2019measure}, DROP \citep{dua2019drop}, and BBH \citep{suzgun2023bbh}.
    \item \textbf{Math}: AIME \citep{maa_aime} and MATH \citep{hendrycks2021math}.
    \item \textbf{Composite}: LiveBench \citep{white2024livebench}, SuperGPQA \citep{mapteam2025supergpqa}, AGIEval \citep{zhong2023agieval}, MMLU-Pro \citep{wang2024mmlupro}, C-Eval \citep{huang2023ceval}, and MMLU \citep{hendrycks2021mmlu}.
    \item \textbf{Long-Context}: RULER \citep{hsieh2024ruler}.
\end{itemize}

\begin{table}[h]
\centering
\caption{Benchmark results. Higher is better. Deltas are score differences versus the dense companion.}
\label{tab:autoeval_benchmarks_32k}
\small
\setlength{\tabcolsep}{0.8pt}
\renewcommand{\arraystretch}{0.92}
\begin{minipage}[t]{0.49\linewidth}
\raggedright
\begin{tabular*}{\linewidth}{@{\extracolsep{\fill}}>{\raggedright\arraybackslash}p{0.38\linewidth}>{\centering\arraybackslash}p{0.12\linewidth}>{\centering\arraybackslash}p{0.205\linewidth}>{\centering\arraybackslash}p{0.205\linewidth}@{}}
\toprule
\textbf{Benchmark} & \textbf{Dense} & \textbf{\densefourx{}} & \textbf{Sparse} \\
\midrule
\multicolumn{4}{@{}l@{}}{\textbf{\textit{Factual QA}}} \\
\midrule
SimpleQA & 5.0 & 3.9\negdelta{1.1} & 3.8\negdelta{1.2} \\
Chinese SimpleQA & 29.7 & 22.4\negdelta{7.3} & 27.5\negdelta{2.2} \\
\midrule
\multicolumn{4}{@{}l@{}}{\textbf{\textit{Code}}} \\
\midrule
MBPP & 72.6 & 66.2\negdelta{6.4} & 72.6\posdelta{0.0} \\
MBPP+ & 67.2 & 62.2\negdelta{5.0} & 65.1\negdelta{2.1} \\
MultiPL-E & 53.4 & 46.8\negdelta{6.6} & 54.1\posdelta{0.7} \\
MBXP & 67.2 & 62.2\negdelta{5.0} & 66.8\negdelta{0.4} \\
LiveCodeBench & 34.4 & 23.9\negdelta{10.5} & 26.9\negdelta{7.5} \\
\midrule
\multicolumn{4}{@{}l@{}}{\textbf{\textit{Reasoning}}} \\
\midrule
KOR-Bench & 43.8 & 36.7\negdelta{7.0} & 43.0\negdelta{0.7} \\
ZebraLogic & 12.5 & 10.8\negdelta{1.7} & 13.3\posdelta{0.8} \\
ProcBench & 10.1 & 11.1\posdelta{1.0} & 11.3\posdelta{1.2} \\
ARC-AGI & 8.9 & 6.1\negdelta{2.8} & 6.4\negdelta{2.5} \\
DROP & 69.2 & 63.8\negdelta{5.5} & 67.9\negdelta{1.4} \\
BBH & 71.9 & 70.8\negdelta{1.1} & 77.8\posdelta{6.0} \\
\bottomrule
\end{tabular*}
\end{minipage}\hfill
\begin{minipage}[t]{0.49\linewidth}
\raggedright
\begin{tabular*}{\linewidth}{@{\extracolsep{\fill}}>{\raggedright\arraybackslash}p{0.38\linewidth}>{\centering\arraybackslash}p{0.12\linewidth}>{\centering\arraybackslash}p{0.205\linewidth}>{\centering\arraybackslash}p{0.205\linewidth}@{}}
\toprule
\textbf{Benchmark} & \textbf{Dense} & \textbf{\densefourx{}} & \textbf{Sparse} \\
\midrule
\multicolumn{4}{@{}l@{}}{\textbf{\textit{Math}}} \\
\midrule
AIME 2025 & 2.0 & 2.3\posdelta{0.3} & 0.7\negdelta{1.3} \\
AIME 2024 & 6.3 & 3.0\negdelta{3.3} & 1.3\negdelta{5.0} \\
MATH & 57.7 & 51.9\negdelta{5.7} & 57.1\negdelta{0.6} \\
\midrule
\multicolumn{4}{@{}l@{}}{\textbf{\textit{Composite}}} \\
\midrule
LiveBench & 30.5 & 26.5\negdelta{4.0} & 28.2\negdelta{2.3} \\
SuperGPQA & 20.0 & 23.2\posdelta{3.2} & 24.9\posdelta{4.9} \\
AGIEval & 64.6 & 58.9\negdelta{5.7} & 64.4\negdelta{0.3} \\
MMLU-Pro & 50.0 & 42.4\negdelta{7.7} & 48.8\negdelta{1.3} \\
C-Eval & 83.4 & 80.6\negdelta{2.8} & 80.8\negdelta{2.6} \\
MMLU & 77.0 & 75.4\negdelta{1.7} & 77.1\posdelta{0.1} \\
\midrule
\multicolumn{4}{@{}l@{}}{\textbf{\textit{Long Context}}} \\
\midrule
RULER 4K & 93.6 & 94.3\posdelta{0.7} & 93.8\posdelta{0.2} \\
RULER 8K & 92.6 & 92.1\negdelta{0.5} & 92.5\negdelta{0.1} \\
RULER 16K & 89.3 & 88.6\negdelta{0.6} & 88.6\negdelta{0.7} \\
RULER 32K & 79.2 & 73.5\negdelta{5.8} & 80.4\posdelta{1.1} \\
\bottomrule
\end{tabular*}
\end{minipage}
\end{table}

Across the results, our predictor-gated sparsity is usually closer to the dense companion than \densefourx{}. The sparse model matches or slightly exceeds dense on MBPP, MultiPL-E and MMLU, etc. It also substantially reduces the \densefourx{} degradation on Chinese SimpleQA, MBPP+, MBXP and KOR-Bench, etc. The sparse model performs better than \densefourx{} at retaining dense quality.

\begin{figure}[h]
    \centering
    \includegraphics[width=\linewidth]{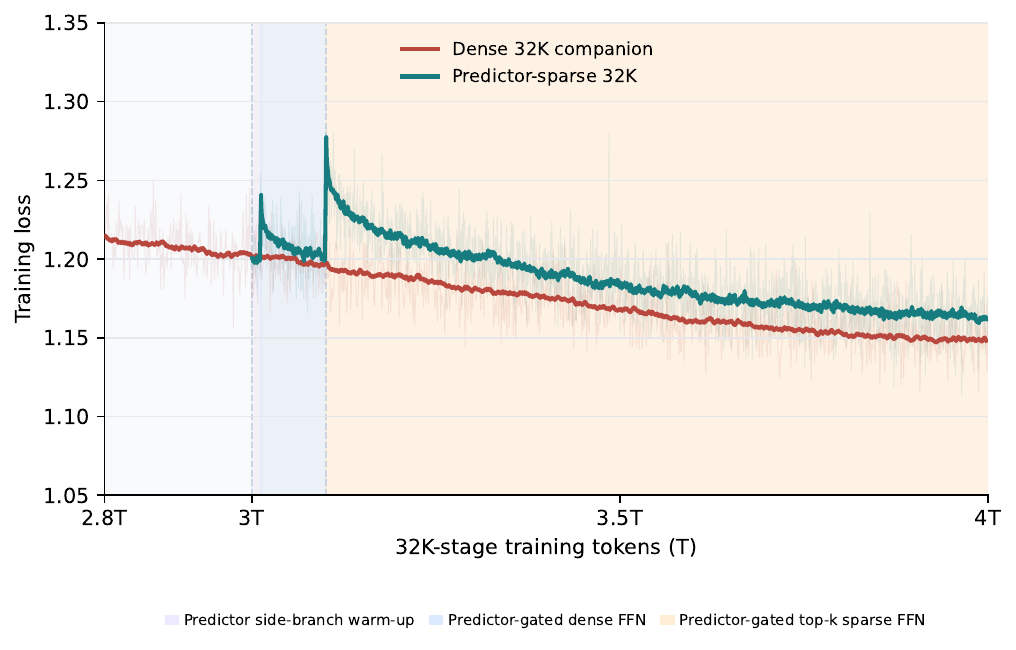}
    \caption{Training loss around the sparse transition in the 32K stage. Colored bands denote predictor side-branch warm-up, predictor-gated dense FFN, and predictor-gated top-$k$ sparse FFN.}
    \label{fig:training_loss_32k}
\end{figure}

\paragraph{Loss Curves.}
Figure~\ref{fig:training_loss_32k} reports the training-loss trajectory around the sparse transition in the 32K stage. The dense model is first trained through the 8K stage and is then continued for a 4T-token 32K stage; the figure zooms into the late 32K continuation where sparse adaptation is introduced. The colored bands correspond to the sparse-transition phases in the training recipe: predictor side-branch warm-up, predictor-gated dense FFN, and predictor-gated top-$k$ sparse FFN.

\section{Training Lessons and Failure Modes}

\subsection{Channel Balancing}

MoE models often use expert-balancing mechanisms to avoid routing collapse. Since our FFN channels behave like fine-grained experts, it is natural to ask whether an expert-balance bias can be applied at the channel level. We evaluate a channel-level routing bias inspired by MoE expert balancing. In our setting, this does not improve the final training outcome, so the main recipe does not adopt a channel-level balancing strategy.

\paragraph{Balance-bias variants.}
For a token $t$ and FFN channel $c$, let $r_{t,c}$ be the predictor logit and $s_{t,c}=\sigma(r_{t,c})$ be the predictor score used by the hard top-$k$ gate. Channels are partitioned into banks $\mathcal{B}_j$ of size 64, and 4x sparsity keeps $k=16$ channels in each bank. A balance-bias vector $b$ is not optimized by backpropagation; it is an online routing correction added only for selection:
\begin{align}
\mathcal{S}_{t,j}
&= \operatorname{topk}_{c \in \mathcal{B}_j}(s_{t,c}+b_c, k), \\
g_{t,c}
&= s_{t,c}\,\mathbf{1}[c \in \mathcal{S}_{t,j}].
\end{align}
The FFN intermediate activation is then multiplied by $g_{t,c}$ as in the normal predictor-sparse path. After accumulating one training step and all-reducing token counts across data-parallel ranks, we compute the selected-token count
\begin{align}
n_c = \sum_t \mathbf{1}[c \in \mathcal{S}_{t,j}], \qquad
\bar n = \frac{1}{C}\sum_{c=1}^{C} n_c ,
\end{align}
where $C$ is the number of local FFN channels in the tensor-parallel shard. We evaluated two update rules. Average balance updates every channel toward the mean usage,
\begin{align}
b_c \leftarrow b_c + \alpha\,\operatorname{sign}(\bar n - n_c).
\end{align}
The 0.05-threshold balance updates only severely under-used channels:
\begin{align}
b_c \leftarrow b_c
+ \alpha\,\mathbf{1}[n_c < 0.05\bar n]\,\operatorname{sign}(\bar n - n_c).
\end{align}
The no-balance baseline fixes $b_c=0$ and applies no online update. In the ablation below, $\alpha$ is the same small step size across balance variants.

\begin{figure}[h]
    \centering
    \includegraphics[width=0.92\linewidth]{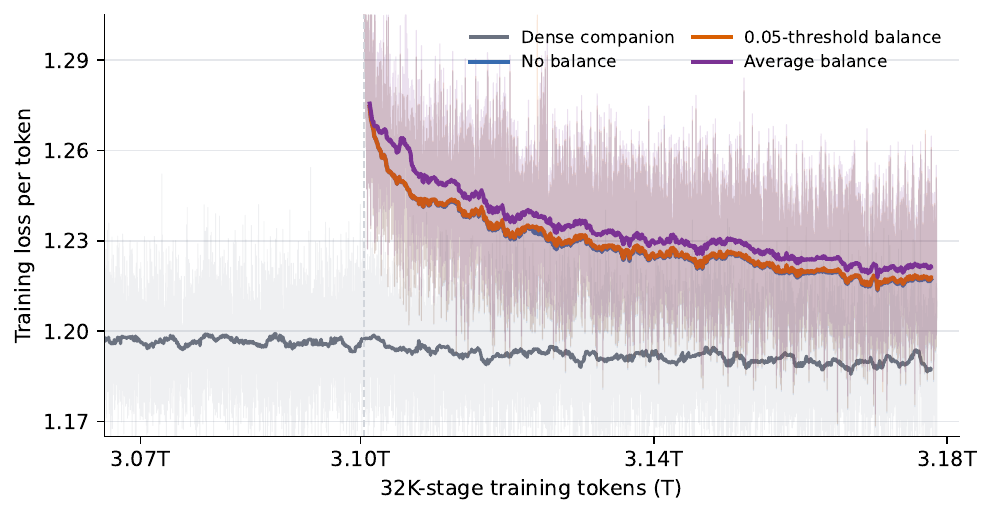}
    \caption{Channel-balance ablations during 32K continual training.}
    \label{fig:balance_mode_loss}
\end{figure}

Figure~\ref{fig:balance_mode_loss} shows the training-loss comparison that motivates this decision. The ablation runs in the 32K continual-training window, where the predictor-sparse model is initialized from a dense checkpoint. We compare no channel-balance bias, average balance, and 0.05-threshold balance.

In this ablation window, average balance is consistently worse, and the 0.05-threshold balance does not improve over the no-balance branch. Therefore, explicitly forcing channel balance is not beneficial. This is a useful negative result. Channel-level routing differs from expert-level routing: channels are much finer-grained, grouped into fixed banks, and coupled through the same FFN matrices. A balance algorithm that works for complete experts may over-constrain or destabilize channel-level selection. We note that this observation is specific to continual training from a dense checkpoint; it does not rule out balance mechanisms during training from scratch, where the router and FFN channels can co-adapt from initialization.

\subsection{Layer-local Predictor Failure on RULER-CWE}
\label{sec:ruler_failure}

\begin{figure}[h]
    \centering
    \includegraphics[width=0.92\linewidth]{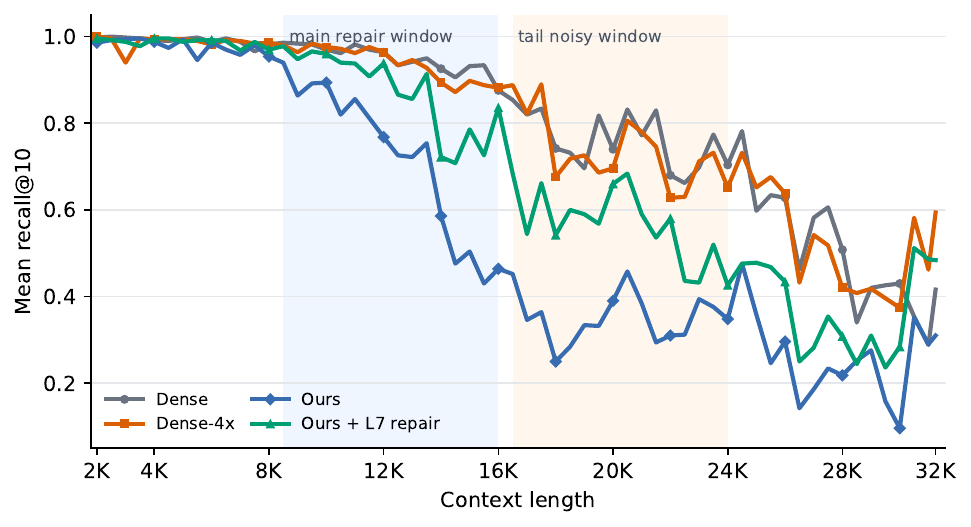}
    \caption{RULER-CWE average recall@10 as context length increases.}
    \label{fig:cwe_length_recall}
\end{figure}

Most short-context and general benchmarks appear close to the dense companion, but RULER-CWE exposes a sharper long-context failure mode. Figure~\ref{fig:cwe_length_recall} presents a fine-grained length sweep with 512-token increments, averaged over 50 independent runs with different seeds. The sparse model is healthy up to 8K but develops an early cliff around 12K--16K, while the dense model remains much more stable at those lengths. This failure is not a global collapse of the sparse model. On short-context reasoning, the sparse model remains close to dense: MMLU accuracy is 0.7404 for sparse v.s. 0.7439 for dense, and ARC-Challenge is 0.9036 for sparse v.s. 0.9181 for dense.

To diagnose the failure, we use a dense fallback intervention that bypasses the sparse predictor at selected FFN layers and token regions. For a target layer, let $m^{\mathrm{sparse}}$ be the sparse FFN intermediate activation and $m^{\mathrm{dense}}$ be the dense FFN intermediate activation. We compute a norm calibration factor
\begin{align}
r_l = \sqrt{\frac{\mathbb{E}\lVert m^{\mathrm{sparse}}_l\rVert_2^2}{\mathbb{E}\lVert m^{\mathrm{dense}}_l\rVert_2^2}},
\end{align}
and replace the selected sparse intermediate activation with $r_l m^{\mathrm{dense}}_l$.

The main positive result is localized to the 7-th layer. In the sweep, layer-7 repair improves the 8704--12288 range average from 0.856 to 0.947 and the 12800--16384 range average from 0.583 to 0.802. This result shows that the RULER-CWE cliff can be traced to a narrow layer-local routing failure instead of a whole-model collapse. Appendix~\ref{app:ruler_repair} gives the detailed phase and token-region localization, as well as the observed trade-offs with short-context reasoning benchmarks.

\section{Conclusion}

We present a channel-level dense-to-sparse continual-training recipe for an 8B-scale language model. The central idea is to reinterpret FFN intermediate channels as fine-grained experts and to use a lightweight predictor as a token-wise router over those channels. The recipe imposes 4x bank-wise channel sparsity, trains the predictor on the main language-modeling path, and converts a dense 8K-to-32K training trajectory into a hardware-oriented sparse FFN model. Current benchmark results show that the predictor-gated sparse model is often much closer to the dense companion than a naive \densefourx{} baseline. The training analysis also exposes two useful failure modes: channel-level balance bias does not transfer from expert-level MoE training, and RULER-CWE can reveal layer-local long-context predictor failures that broad short-context benchmarks do not expose. The single-layer repair analysis shows that part of this long-context failure is localizable and partially recoverable.

\bibliographystyle{unsrtnat}
\bibliography{references}

\clearpage
\appendix
\section*{Appendix}
\section{Predictor Top-k Ablation}
\label{app:predictor_topk_ablation}

This section records the diagnostic ablation that motivates the predictor-before-top-$k$ sparse-routing order. This experiment is run on a 200M-parameter small diagnostic model. It is not a final quality claim for the 8B sparse model. The ablation compares several ways to impose sparse FFN routing after a common continual-training checkpoint:
\begin{itemize}
    \item \textbf{Direct top-$k$}: Selects active channels directly from native FFN channel scores.
    \item \textbf{Direct top-$k$ with compensation}: Keeps the same selected channels but rescales the post-mask activation.
    \item \textbf{Predictor-before-top-$k$ (adopted recipe)}: Places the low-rank predictor before bank-wise hard top-$k$ selection.
\end{itemize}

\begin{figure}[h]
    \centering
    \includegraphics[width=0.92\linewidth]{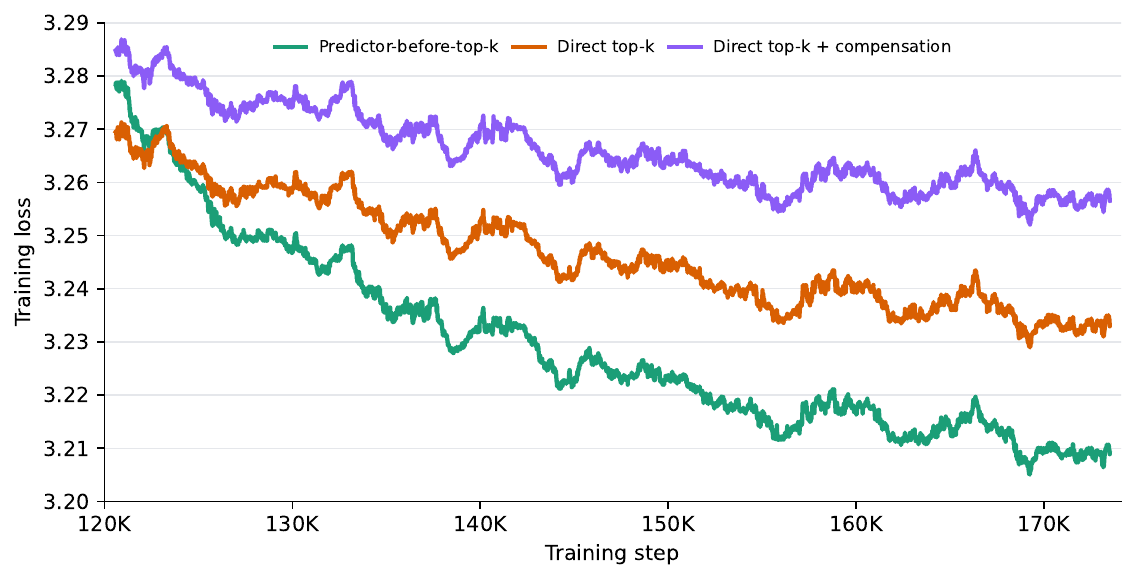}
    \caption{Continual-training ablation comparing predictor before top-$k$ sparse routing with direct top-$k$ masking variants.}
    \label{fig:direct_topk_ablation}
\end{figure}

Figure~\ref{fig:direct_topk_ablation} shows the corresponding training loss comparison. The result supports using a trainable predictor before the hard top-$k$ operator. Direct top-$k$ masking is simpler, but it gives worse loss performance.

\clearpage
\section{RULER-CWE Repair Analysis}
\label{app:ruler_repair}

This section records the details behind the layer-7 repair result in Section~\ref{sec:ruler_failure}. The repair replaces selected sparse FFN intermediate activations with norm-calibrated dense activations at one layer and then measures which task behavior is recovered or damaged.

\textbf{Prompt Prefill Repair is Sufficient for the CWE Recovery.}
We first separate the layer-7 repair by execution phase. In the all-call condition, the calibrated dense fallback is applied whenever the target layer is evaluated. In the prompt-processing condition, it is applied only while consuming the input prompt. In the decoding-only condition, it is applied only after generation begins. Table~\ref{tab:ruler_phase_repair} reports results averaged over 50 different random seeds, with context length increasing in 512-token steps.

\begin{table}[h]
\centering
\caption{Phase localization for layer-7 repair on RULER-CWE. We report the recall@10.}
\label{tab:ruler_phase_repair}
\small
\begin{tabular}{lccccc}
\toprule
Context range & Dense & Sparse & Repair All & Repair Prefill & Repair Decode \\
\midrule
2048--8192 & 0.991 & 0.978 & 0.989 & 0.987 & 0.981 \\
8704--12288 & 0.975 & 0.856 & 0.944 & 0.947 & 0.878 \\
12800--16384 & 0.925 & 0.583 & 0.817 & 0.802 & 0.604 \\
16896--24576 & 0.762 & 0.352 & 0.575 & 0.566 & 0.371 \\
25088--32000 & 0.498 & 0.259 & 0.277 & 0.374 & 0.196 \\
\bottomrule
\end{tabular}
\end{table}

The main recovery therefore comes from how the model processes the prompt, not from changing the tokens produced during autoregressive decoding.

\textbf{Effective Token Region: Final Question-and-Answer Suffix.}
We then restrict the same layer-7 prompt-processing repair to different final prompt regions. Table~\ref{tab:ruler_suffix_repair} shows that the strongest localized repair comes from patching the suffix where the prompt asks the question and begins the answer. Repairing the final 64 tokens and repairing the question-and-answer suffix produce nearly identical CWE recovery, while excluding that suffix removes most of the gain.

\begin{table}[h]
\centering
\caption{Token-region localization for layer-7 prompt-processing repair.}
\label{tab:ruler_suffix_repair}
\small
\setlength{\tabcolsep}{4pt}
\renewcommand{\arraystretch}{1.08}
\begin{tabular}{lcc}
\toprule
Repair condition & CWE recall@10 & MMLU \\
\midrule
No repair & 0.133 & 0.702 \\
Final 64 tokens & 0.933 & 0.628 \\
Question-and-answer suffix & 0.933 & 0.632 \\
Final 64 tokens excluding Q\&A suffix & 0.267 & 0.702 \\
Answer-prefix only & 0.650 & 0.698 \\
\bottomrule
\end{tabular}
\end{table}

This points to a readout-query failure. The sparse model appears to process enough evidence that a localized suffix repair can recover the common-word answer, but the predictor routing at the final question-and-answer suffix is not reliable for this long-CWE prompt.

\textbf{Repair is not a General Long-Context Fix.}
The same intervention does not repair Needle-In-A-Haystack (NIAH)~\citep{kamradt2023needle}, where the model must retrieve and reason over a needle embedded in long context. Table~\ref{tab:niah_repair} shows that final-suffix repairs do not close the dense-sparse gap.

\begin{table}[h]
\centering
\caption{NIAH recall under the same layer-7 repair interventions.}
\label{tab:niah_repair}
\small
\setlength{\tabcolsep}{4pt}
\renewcommand{\arraystretch}{1.08}
\begin{tabular}{lc}
\toprule
Condition & NIAH recall \\
\midrule
Dense companion & 0.8636 \\
Sparse baseline & 0.7136 \\
Final 64-token repair & 0.7111 \\
Question-and-answer repair & 0.7111 \\
Answer-prefix only & 0.7161 \\
\bottomrule
\end{tabular}
\end{table}

This negative result suggests that the CWE repair restores a small amount of low-level readout functionality rather than a general long-context understanding mechanism. Tasks such as NIAH appear to require distributed retrieval and integration behavior that cannot be recovered by patching one layer and one final suffix region.

\begin{figure}[h]
    \centering
    \includegraphics[width=0.92\linewidth]{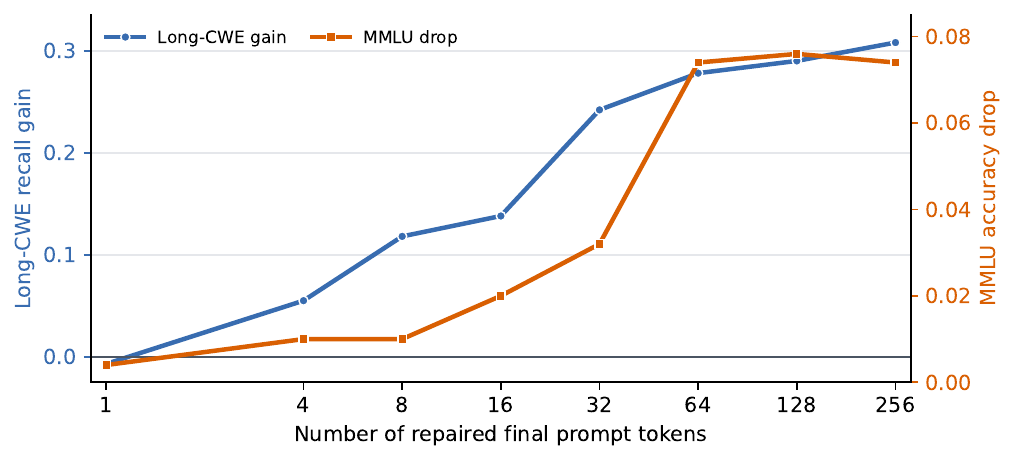}
    \caption{Layer-7 final-token repair shows a trade-off between long-CWE recovery and MMLU.}
    \label{fig:ruler_repair_tradeoff}
\end{figure}

\textbf{Repair Trades off with MMLU.}
The repair also trades off against short-context reasoning. Figure~\ref{fig:ruler_repair_tradeoff} varies the number of repaired final prompt tokens. As the repaired window grows, the average long-CWE gain increases, while MMLU accuracy decreases relative to the sparse baseline.

One plausible explanation is that the same layer-7 suffix routing subspace that is fragile for long-CWE readout is also useful for ordinary semantic selection in short benchmarks. Replacing it with dense activations can recover the narrow CWE behavior while disrupting the sparse model's learned predictor routing on MMLU.

\end{document}